\documentclass[11pt]{article}
\usepackage[preprint]{acl}

\usepackage{times}
\usepackage{latexsym}
\usepackage[table]{xcolor}
\usepackage[normalem]{ulem}
\usepackage{arydshln}
\usepackage[T1]{fontenc}
\usepackage{float}
\usepackage{xspace}
\newcommand{\tvext}[1]{\text{#1}}
\usepackage[utf8]{inputenc}
\usepackage[most]{tcolorbox}
\usepackage{microtype}
\usepackage{multirow}
\usepackage{tcolorbox}
\usepackage{graphicx}
\usepackage{tcolorbox}
\usepackage{placeins}
\tcbuselibrary{skins,breakable}

\usepackage{inconsolata}
\usepackage{pgfplots}
\pgfplotsset{compat=1.17}
\usepackage{tikz}
\usetikzlibrary{patterns}
\usepackage{graphicx}
\usepackage{booktabs}
\usetikzlibrary{shapes, arrows.meta, positioning, calc, backgrounds, fit, shadows}
\usepackage{amsmath}
\usepackage{amssymb}
\usepackage{enumitem}
\usetikzlibrary{shapes, arrows.meta, positioning, calc, backgrounds, fit, shadows.blur, patterns, decorations.pathreplacing, fadings}

\definecolor{colorA}{RGB}{235, 245, 255}
\definecolor{colorB}{RGB}{255, 240, 230}
\definecolor{colorC}{RGB}{235, 250, 235}
\definecolor{colorD}{RGB}{245, 235, 250}
\definecolor{darkblue}{RGB}{20, 50, 100}
\definecolor{stopred}{RGB}{220, 50, 50}
\newcommand\method{\text{NEAT}\xspace}

\newtcolorbox{promptbox}[1]{
  enhanced,
  colback=white,
  colframe=black,
  boxrule=0.8pt,
  arc=2pt,
  left=6pt,
  right=6pt,
  top=6pt,
  bottom=6pt,
  title=#1,
  colbacktitle=black,
  coltitle=white,
  fonttitle=\bfseries,
  titlerule=0pt,              
  boxed title style={boxrule=0pt}
}

\title{NEAT: Neuron-Based Early Exit for Large Reasoning Models}

\author{
  Kang Liu, Yongkang Liu, Xiaocui Yang, Peidong Wang, \\
  \textbf{Wen Zhang, Shi Feng\textsuperscript{\textdagger}, Yifei Zhang, Daling Wang\textsuperscript{\textdagger}} \\
  Northeastern University, China \\
  \texttt{lk\_stu\_neu@163.com, \{fengshi, wangdaling\}@cse.neu.edu.cn}
}
\begin{document}
\maketitle
{
  \renewcommand{\thefootnote}{\fnsymbol{footnote}}
  \footnotetext[2]{Corresponding authors.}
}
\begin{abstract}

Large Reasoning Models (LRMs) often suffer from \emph{overthinking}, a phenomenon in which redundant reasoning steps are generated after a correct solution has already been reached. Existing early reasoning exit methods primarily rely on output-level heuristics or trained probing models to skip redundant reasoning steps, thereby mitigating overthinking. However, these approaches typically require additional rollout computation or externally labeled datasets. In this paper, we propose \textbf{NEAT}, a \textbf{N}euron-based \textbf{E}arly re\textbf{A}soning exi\textbf{T} framework that monitors neuron-level activation dynamics to enable training-free early exits, without introducing additional test-time computation. NEAT identifies exit-associated neurons and tracks their activation patterns during reasoning to dynamically trigger early exit or suppress reflection, thereby reducing unnecessary reasoning while preserving solution quality. Experiments on four reasoning benchmarks across six models with different scales and architectures show that, for each model, NEAT achieves an average token reduction of 22\% to 28\% when averaged over the four benchmarks, while maintaining accuracy.

\end{abstract}

\section{Introduction}

Large Reasoning Models (LRMs), such as OpenAI o1~\cite{o1} and DeepSeek-R1~\cite{guo2025deepseek}, have achieved substantial performance gains on complex reasoning tasks, including mathematical problem solving and competitive programming—by scaling sequential test-time computation~\cite{snell2024scaling}. 
However, LRMs often perform more reasoning steps than necessary, continuing generation even after correct answers have been reached. This behavior, referred to as \emph{overthinking}~\cite{sui2025stop}, leads to excessive computational consumption. 
Beyond increased inference latency, overthinking also raises the risk of hallucination and error accumulation in the generation tail~\cite{wu2025more,yu2024interpreting}. Several prior works attempt to mitigate this issue by training models to reason more efficiently through supervised fine-tuning (SFT)~\cite{zhao2025let} or reinforcement learning (RL)~\cite{zhang2025adaptthink,lou2025adacot}. 
In contrast, a more direct and cost-effective alternative is to identify the moment when reasoning has effectively completed and exit the reasoning process early, thereby skipping redundant steps.

\begin{figure}[!t]
    \centering
    \includegraphics[width=0.5\textwidth]{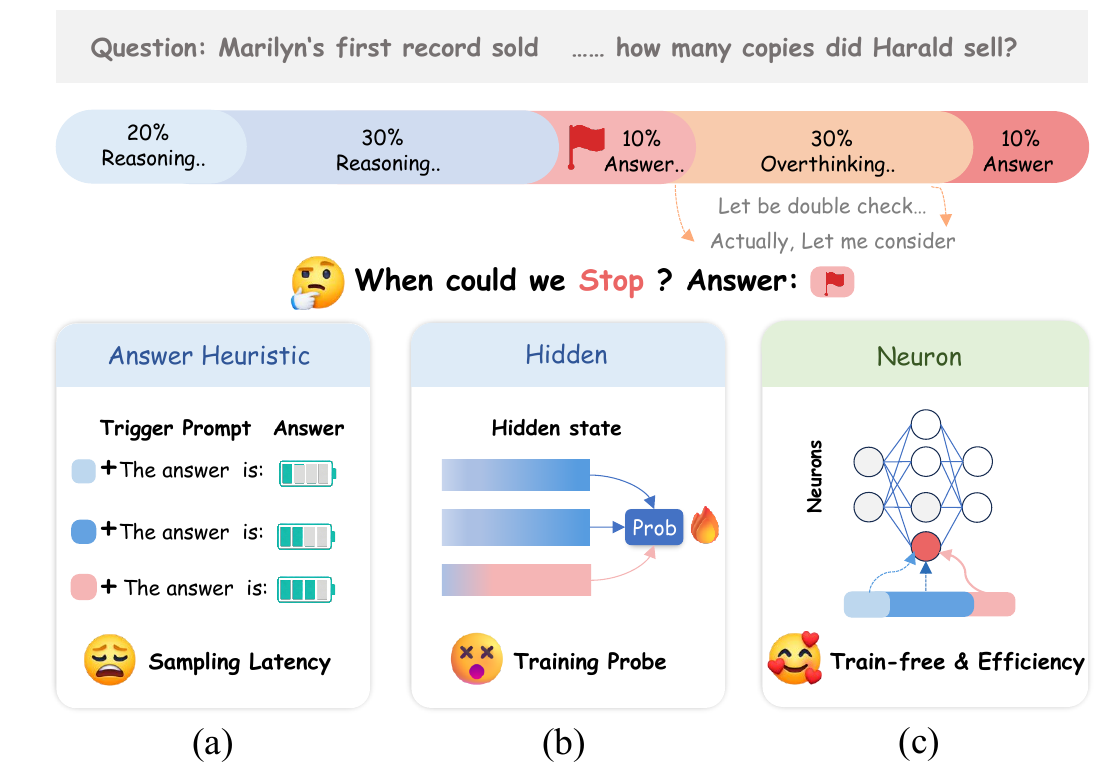}
    \caption{Comparison of different early reasoning exit methods. (a) Output-based Answer Heuristic methods (b) Probe-based method. (c) Our method NEAT.}
    \label{fig:intro}
\end{figure}

A widely explored line of work estimates reasoning completion by evaluating the convergence of intermediate outputs, such as answer consistency~\cite{fu2024efficiently,huang2025efficient} or confidence-based criteria~\cite{yang2025dynamic,liu2025answer}. 
While effective, these approaches typically require repeated rollouts, resulting in increased parallel test-time computation and inference latency, as illustrated in Figure~\ref{fig:intro}(a). Another line of work explores information encoded in latent hidden states during reasoning to provide early exit signals and avoid additional sampling latency. 
Although promising, existing methods~\cite{zhang2025reasoning,liu2025answer,eisenstadt2025overclocking} generally rely on externally supervised probes trained on hidden states, as shown in Figure~\ref{fig:intro}(b).

In this paper, we further explore the potential of internal latent states from a more fine-grained, neuron-level perspective to provide efficient early-exit signals, and propose \textbf{\method} (\textbf{N}euron-based \textbf{E}arly re\textbf{A}soning exi\textbf{T}), a training-free framework that determines whether to exit or continue reasoning by leveraging neuron-level activation dynamics, without requiring external supervision or parallel inference-time computation.
Specifically, our method first identifies a sparse set of \textit{exit-associated neurons} whose activation patterns are tightly coupled with natural reasoning termination. 
These neurons are selected based on their causal contribution to termination-related token predictions and exhibit activation values whose temporal mass is concentrated toward the end of the reasoning process. During inference, \textsc{\method} monitors the real-time activation dynamics of these neurons to assess whether the model has internally converged. To mitigate the risk of premature termination, we incorporate a dynamic intervention strategy that leverages the strength of observed neuron activation patterns as indicators of reasoning completion. When the signal is strong, the model performs early exit, whereas when the signal is relatively weak, it suppresses reflection tokens and allows reasoning to continue. Experiments across multiple reasoning benchmarks and model architectures show that \textsc{\method} reduces redundant token generation while maintaining final-answer accuracy.

Our contributions are summarized as follows:
\begin{itemize}

    \item  We demonstrate that neuron-level activation dynamics provide reliable internal signals for early exit, offering a fine-grained alternative to output-level answer-convergence methods.
    
    \item   We propose \textbf{\method}, a training-free early reasoning-exit framework that monitors neuron-level activation dynamics and performs two inference-time interventions, early exit and reflection suppression, to control the progression of reasoning, without requiring external supervision or parallel rollouts.
    
    \item  Experiments across various model architectures and reasoning benchmarks demonstrate that NEAT reduces the average token count by 22.0\% to 28.2\% while maintaining accuracy.
\end{itemize}

\section{Related Work}

\paragraph{Early Reasoning Exit.}
Early reasoning exit methods aim to precisely identify the point at which a reasoning model has gathered sufficient information to produce a correct answer, allowing the model to skip redundant steps and thus mitigate overthinking during inference. Most existing studies primarily rely on output-based heuristic methods. For example, Dynasor~\cite{fu2024efficiently} periodically probes intermediate answers at fixed token intervals and triggers an early exit when multiple consecutive outputs are consistent. However, probing answer convergence at fixed steps incurs additional overhead and lacks flexibility. Alternatively, DEER \cite{yang2025dynamic} alleviates this issue by triggering exits only at reflection keywords (e.g., Wait), guided by answer confidence. However, direct exiting may hurt performance. Thus, CGRS~\cite{huang2025efficient} proposes to only suppress reflection triggers when the model is confident in its current response.  Another line of work leverages internal hidden states to provide early-exit signals without additional sampling overhead. For example, TPV~\cite{eisenstadt2025overclocking} trains probes to encode internal reasoning progress and manipulates these probe signals during inference to skip redundant reasoning steps. Similarly, \citet{zhang2025reasoning} trains probes on hidden states to predict answer correctness. While such methods avoid additional computation during reasoning, they require extra supervised data to train probes and still exhibit a notable performance gap compared to mature output-based approaches~\cite{yang2025dynamic}.

\paragraph{Neurons in LLMs.}
Neurons are the most fundamental computational units in LLMs. Prior work has identified individual neurons that are closely associated with diverse model capabilities and behaviors~\cite{tang2024language,shi2024ircan,chen2024journey}. 
For reasoning tasks, studies such as~\citet{yu2024interpreting,rai2024investigation} show that only a small subset of neurons is selectively activated, and that these neurons play an important role in enabling arithmetic reasoning. Moreover, \citet{yu2025back} finds that failures in multi-step reasoning can often be attributed to neurons being activated at incorrect positions during the reasoning process, rather than to a lack of relevant knowledge. Beyond deepening the analysis and understanding of model behavior, identifying such neurons has also enabled several practical applications~\cite{gurgurov2025language,yu2025locate}. For example, \citet{tang2025enhancing} improves the quality of chain-of-thought reasoning by directly stimulating neurons identified as critical for reasoning. EELo-CoT~\cite{zhao2025activation} further designs an activation control module that efficiently elicits long chain-of-thought behavior for base models. 
In this work, we further explore neuron-level signals as a real-time control mechanism for early reasoning exit, thereby enabling more efficient inference.

\begin{figure*}[!h]
    \centering
    \includegraphics[width=1.\textwidth]{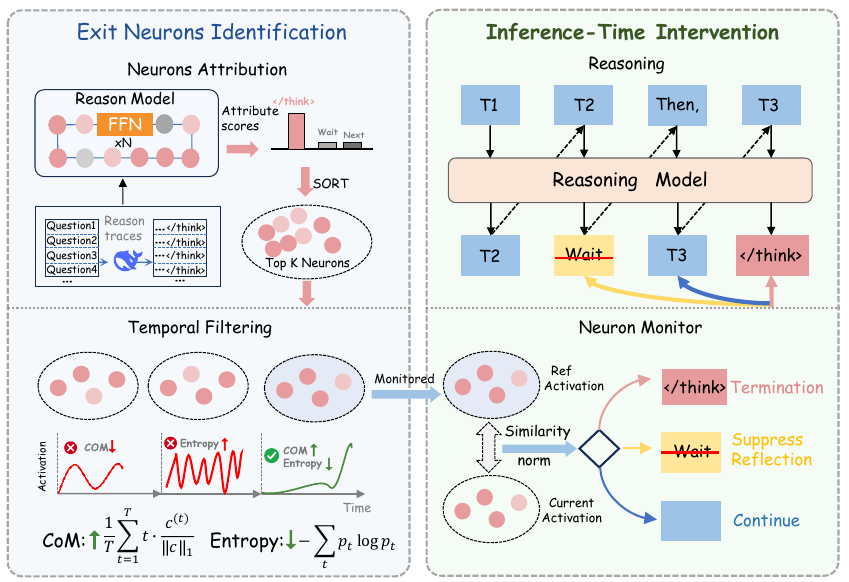}
    \caption{Overview of the proposed \method framework. \textbf{Left:} During calibration, we identify exit-associated neurons by computing attribution scores at the termination timestep and filtering based on temporal activation patterns. \textbf{Right:} During inference, we monitor the activation dynamics of the identified neurons and apply dynamic intervention when the pattern matches the reference regime.}
    \label{fig:framework}
\end{figure*}

\section{Methodology}
\label{sec:method}

We propose \textbf{\method}, an early reasoning exit framework  by monitoring model's internal neuron activation dynamics.
As illustrated in Figure~\ref{fig:framework}, our approach mainly consists of two stages:
(1) \textbf{Exit Neurons Identification} (\S\ref{sec:discovery}), which identifies a sparse set of neurons whose activations are strongly associated with reasoning exit
(2) \textbf{Inference-Time Intervention} (\S\ref{sec:inference}),which monitors these neurons during decoding and intervenes in the reasoning process in real time based on their activation patterns.

\subsection{Exit Neurons Identification}
\label{sec:discovery}

Large Reasoning Models (LRMs) generate a reasoning trace $\mathbf{R} = [r^{(1)}, r^{(2)}, \dots, r^{(T)}]$ before producing the final answer, where each $r_t$ represents a reasoning step.
The completion of reasoning is signaled by a special termination token $w_{\text{end}}$ (e.g., \texttt{</think>}), which marks the transition from reasoning to answer generation. Our goal is to identify neurons whose activations provide early internal signals of reasoning termination before the explicit generation of $w_{\text{end}}$.
We achieve this by attributing neurons to termination token prediction and selecting those that exhibit consistent activation toward the end of the reasoning trace.


\paragraph{Neuron Attribution.}
In the inference pass in decoder-only LLMs, for a given input sequence, each layer output $\mathbf{h}_i^l$ (layer $l$, token position $i$) is a sum of the previous layer's output $\mathbf{h}_i^{l-1}$, the attention output $\mathbf{A}_i^l$, and the FFN output $\mathbf{F}_i^l$:
\begin{equation}
\mathbf{h}_i^l = \mathbf{h}_i^{l-1} + \mathbf{A}_i^l + \mathbf{F}_i^l.
\end{equation}
The FFN output $\mathbf{F}_i^l$ is calculated by a non-linear $\sigma$ on two MLPs $W_{fc1}^l \in \mathbb{R}^{N \times d}$ and $W_{fc2}^l \in \mathbb{R}^{d \times N}$:
\begin{equation}
\mathbf{F}_i^l = W_{fc2}^l \sigma(W_{fc1}^l (\mathbf{h}_i^{l-1} + \mathbf{A}_i^l)).
\end{equation}
Following~\cite{geva2021transformer}, the FFN layer output $\mathbf{F}_i^l$ can be represented as a weighted sum over neuron subvalues:
\begin{equation}
\mathbf{F}_i^l = \sum_{k=1}^{N} c_{i,k}^l \cdot fc2_k^l,
\end{equation}
where $fc2_k^l$ denotes the $k$-th column of $W_{fc2}^l$, and $c_{i,k}^l$ is the activation of the $k$-th neuron:
\begin{equation}
c_{i,k}^l = \sigma(fc1_k^l \cdot (\mathbf{h}_i^{l-1} + \mathbf{A}_i^l)),
\end{equation}
where $fc1_k^l$ denotes the $k$-th row of $W_{fc1}^l$.

To quantify the importance of each neuron for generating the termination token (e.g., \texttt{</think>}), we adopt the log probability increase method of~\cite{yu2024neuron}. For a neuron in the $l$-th FFN layer, denoted as $v^l$, its importance score is defined as the increase in log probability of the target token when $v^l$ is added to the residual stream $\mathbf{A}^l + \mathbf{h}^{l-1}$, compared to the baseline without $v^l$:
\begin{equation}
\label{eq:importance}
\begin{aligned}
\text{Imp}(v^l) &= \log p(w_{\text{end}} | v^l + \mathbf{A}^l + \mathbf{h}^{l-1}) \\
&\quad - \log p(w_{\text{end}} | \mathbf{A}^l + \mathbf{h}^{l-1}).
\end{aligned}
\end{equation}
This approach efficiently identifies neurons whose activations most strongly influence the model's prediction at the termination position. We retain the top-$k$ neurons with the highest positive importance scores as candidates.

\paragraph{Temporal Filtering.}
High attribution at a single decoding step is insufficient to characterize neurons that reliably signal reasoning completion. Thus, we furthermore analyze the activation trajectory $\mathbf{c} = [c^{(1)}, \dots, c^{(T)}]$ of each candidate across the reasoning trace.
We compute two metrics: the \textbf{Relative Center of Mass (CoM)}, which measures where activation concentrates along the trace:
\begin{equation}
\mu_{\text{com}} = \frac{1}{T} \sum_{t=1}^{T} t \cdot \frac{c^{(t)}}{\|\mathbf{c}\|_1},
\end{equation}
and the \textbf{Activation Entropy}, which measures temporal dispersion:
\begin{equation}
H(\mathbf{c}) = -\sum_t p_t \log p_t, \quad p_t = \frac{c^{(t)}}{\|\mathbf{c}\|_1}.
\end{equation}
Neurons with high \textbf{CoM} and low \textbf{Entropy} exhibit focused, late-stage activation and are selected as exit-associated neurons.
We aggregate these metrics over a calibration set $\mathcal{D}_{\text{cal}}$ and define the final neuron set as:
\begin{equation}
\mathcal{S}^* = \{ v \mid \Omega(v) \ge \tau_{\text{cons}} \ \land \ \bar{\mu}_{\text{com}}(v) \ge \tau_{\text{com}} \},
\end{equation}
where $\Omega(v)$ is the frequency of neuron $v$ appearing among top-$k$ candidates across samples, $\bar{\mu}_{\text{com}}(v)$ is the average CoM of neuron over $\mathcal{D}_{\text{cal}}$, $\tau_{\text{cons}}$ is the consistency threshold for filtering neurons that appear frequently, and $\tau_{\tvext{com}}$ is the CoM threshold for selecting neurons with late-stage activation patterns.

\subsection{Inference-Time Intervention}
\label{sec:inference}

During inference, the identified neuron set $\mathcal{S}^*$ is held fixed.
At each decoding step $t$, we extract the activation vector $\mathbf{a}_t \in \mathbb{R}^{|\mathcal{S}^*|}$ corresponding to these neurons and compare it against a reference pattern to determine whether to exit, suppress, or continue reasoning, as illustrated in the right panel of Figure~\ref{fig:framework}.

\paragraph{Activation Pattern Matching}
To assess whether the model has reached an exit-associated state, we compute two alignment measures between the current activation $\mathbf{a}_t$ and a reference pattern $\boldsymbol{\mu}_{\text{ref}}$, which is obtained by averaging the activations of $\mathcal{S}^*$ at the termination timestep over the calibration set $\mathcal{D}_{\text{cal}}$:
\begin{equation}
\rho_t = \text{CosSim}(\mathbf{a}_t, \boldsymbol{\mu}_{\text{ref}}), \quad
\phi_t = \frac{\|\mathbf{a}_t\|_2}{\|\boldsymbol{\mu}_{\text{ref}}\|_2}.
\end{equation}
Here, $\rho_t$ captures the directional similarity between activation patterns, while $\phi_t$ measures the relative activation magnitude.
Together, they indicate how closely the current state resembles the exit regime.

\paragraph{Dynamic Intervention Strategy}
Based on these alignment signals, \textsc{\method} applies one of three actions at each step:

\begin{itemize}[leftmargin=*, itemsep=2pt]
\item \textbf{Termination.}
If $\rho_t > \tau_{\text{sim}}$ and $\phi_t > \tau_{\text{mag}}$, the activation pattern strongly matches the reference regime, indicating that reasoning has effectively completed.
We halt generation and append the termination token \texttt{</think>}.

\item \textbf{Suppress Reflection.}
If $\tau_{\text{sup}} < \rho_t \le \tau_{\text{sim}}$  and $\phi_t > \tau_{\text{mag}}$ but the termination criterion is not met, we suppress a predefined set of reflection tokens by setting their logits to $-\infty$, reducing unnecessary continuation without forcing immediate exit. 

\item \textbf{Continue.}
Otherwise, decoding proceeds normally without intervention.
\end{itemize}

This hierarchical strategy allows \textsc{\method} to aggressively terminate when confidence is high, gently guide the model away from redundant patterns when confidence is moderate, and avoid premature exit when the signal is weak.

\begin{table*}[t]
\centering
\small
\resizebox{\textwidth}{!}{%
\begin{tabular}{lccc ccc ccc ccc cc}
\toprule
\multirow{2}{*}{\textbf{Method}} 
& \multicolumn{3}{c}{\textbf{MATH500}} 
& \multicolumn{3}{c}{\textbf{AMC23}} 
& \multicolumn{3}{c}{\textbf{AIME24}} 
& \multicolumn{3}{c}{\textbf{GPQA-D}} 
& \multicolumn{2}{c}{\textbf{AVG}} \\
\cmidrule(lr){2-4} \cmidrule(lr){5-7} \cmidrule(lr){8-10} \cmidrule(lr){11-13} \cmidrule(lr){14-15}
 & Acc$\uparrow$ & \#Tok$\downarrow$ & LR$\uparrow$
 & Acc$\uparrow$ & \#Tok$\downarrow$ & LR$\uparrow$
 & Acc$\uparrow$ & \#Tok$\downarrow$ & LR$\uparrow$
 & Acc$\uparrow$ & \#Tok$\downarrow$ & LR$\uparrow$
 & Acc$\uparrow$ & LR$\uparrow$ \\
\midrule
\multicolumn{15}{c}{\textit{DeepSeek-R1-Distill-Qwen-7B}} \\
\midrule
Vanilla & 92.2 & 3743 & -- & 87.5 & 5861 & -- & 52.2 & 10662 & -- & 53.0 & 7301 & -- & 71.2 & -- \\
NoThinking & 80.9 & 1173 & 65.4\% & 75.8 & 2499 & 57.4\% & 32.2 & 6680 & 37.3\% & 37.9 & 1312 & 81.8\% & 56.7 & 61.4\% \\\hdashline
TALE & 89.1 & 2657 & 29.0\% & 86.7 & 5107 & 12.9\% & 48.9 & 9727 & 8.8\% & 36.2 & 4938 & 31.3\% & 65.2 & 20.7\% \\
Dynasor & 81.8 & 2070 & 44.7\% & 84.2 & 5201 & 11.3\% & 47.8 & 8334 & 21.8\% & 22.2 & 561 & 92.3\% & 59.0 & \textbf{42.5}\% \\
DEER & 89.6 & 2272 & 39.3\% & 88.3 & 4670 & 20.3\% & 47.8 & 9288 & 12.9\% & 52.5 & 6691 & 8.4\% & 69.6 & 20.2\% \\
CGRS & 92.0 & 2844 & 24.0\% & 88.3 & 3406 & 41.9\% & 52.2 & 7597 & 28.8\% & 53.0 & 5826 & 20.2\% & \underline{71.3} & \underline{28.7}\% \\
\rowcolor{gray!15}\method & 92.2 & 2914 & 22.1\% & 88.3 & 4485 & 23.5\% & 53.3 & 8321 & 22.0\% & 53.0 & 5798 & 20.6\% & \textbf{71.7} & 22.0\% \\
\midrule
\multicolumn{15}{c}{\textit{DeepSeek-R1-Distill-Llama-8B}} \\
\midrule
Vanilla & 85.7 & 4087 & -- & 84.2 & 6374 & -- & 44.9 & 10585 & -- & 49.3 & 7672 & -- & 66.0 & -- \\
NoThinking & 83.3 & 2405 & 41.2\% & 82.5 & 5796 & 9.1\% & 40.0 & 11242 & -6.2\% & 36.2 & 6638 & 13.5\% & 60.5 & 14.4\% \\\hdashline
TALE & 84.0 & 3541 & 13.4\% & 85.0 & 5915 & 7.2\% & 40.0 & 11141 & -5.3\% & 46.1 & 5573 & 27.4\% & 63.8 & 10.7\% \\
Dynasor & 85.0 & 3585 & 12.3\% & 84.2 & 5681 & 10.9\% & 37.7 & 10368 & 2.1\% & 31.8 & 3095 & 59.7\% & 59.7 & \underline{21.2}\% \\
DEER & 82.3 & 2722 & 33.4\% & 80.8 & 5480 & 14.0\% & 42.2 & 9778 & 7.6\% & 41.8 & 7434 & 3.1\% & 61.8 & 14.5\% \\
CGRS & 84.7 & 3254 & 20.4\% & 86.7 & 4899 & 23.1\% & 47.8 & 9536 & 9.9\% & 39.6 & 7221 & 5.9\% & \underline{64.7} & 14.8\% \\
\rowcolor{gray!15}\method & 84.7 & 2934 & 28.2\% & 85.6 & 4945 & 22.4\% & 45.5 & 8596 & 18.8\% & 48.5 & 5841 & 23.9\% & \textbf{66.1} & \textbf{23.3}\% \\
\midrule
\multicolumn{15}{c}{\textit{Qwen3-8B}} \\
\midrule
Vanilla & 94.6 & 5140 & -- & 89.4 & 7876 & -- & 61.1 & 11924 & -- & 57.7 & 9104 & -- & 75.7 & -- \\
NoThinking & 87.1 & 1239 & 75.7\% & 72.5 & 2426 & 69.2\% & 30.0 & 5967 & 50.0\% & 54.2 & 1546 & 83.0\% & 61.0 & 70.0\% \\\hdashline
TALE & 92.3 & 3885 & 24.4\% & 88.3 & 6872 & 12.7\% & 68.9 & 10942 & 8.2\% & 59.1 & 7113 & 21.9\% & \textbf{77.2} & 16.8\% \\
Dynasor & 91.7 & 3841 & 25.3\% & 89.2 & 6457 & 18.0\% & 62.2 & 10174 & 14.7\% & 57.7 & 5965 & 34.5\% & 75.2 & 23.1\% \\
DEER & 88.7 & 1935 & 62.4\% & 79.2 & 3715 & 52.8\% & 45.6 & 7443 & 37.6\% & 59.3 & 7837 & 13.9\% & 68.2 & \textbf{41.7\%} \\
CGRS & 93.3 & 3507 & 31.8\% & 89.2 & 5595 & 29.0\% & 61.1 & 8792 & 26.3\% & 59.8 & 6302 & 30.8\% & \underline{75.8} & \underline{29.5}\% \\
\rowcolor{gray!15}\method & 95.0 & 3906 & 24.0\% & 88.3 & 5820 & 26.1\% & 60.0 & 9067 & 24.0\% & 59.3 & 7485 & 17.8\% & 75.6 & 23.0\% \\
\midrule
\multicolumn{15}{c}{\textit{Qwen3-14B}} \\
\midrule
Vanilla & 94.1 & 4551 & -- & 93.3 & 7190 & -- & 68.9 & 11316 & -- & 64.0 & 7411 & -- & 80.1 & -- \\
NoThinking & 87.0 & 853 & 81.3\% & 77.5 & 1616 & 77.5\% & 27.8 & 3689 & 67.4\% & 56.9 & 1268 & 82.9\% & 62.3 & 77.3\% \\\hdashline
TALE & 93.7 & 3389 & 25.5\% & 92.5 & 5951 & 17.2\% & 71.1 & 10860 & 4.0\% & 63.8 & 6091 & 17.8\% & 80.3 & 16.1\% \\
Dynasor & 84.4 & 3667 & 19.4\% & 90.0 & 6030 & 16.1\% & 65.6 & 9775 & 13.6\% & 64.3 & 5775 & 22.1\% & 76.1 & 17.8\% \\
DEER & 91.7 & 1956 & 57.0\% & 90.8 & 4079 & 43.3\% & 56.7 & 6755 & 40.3\% & 58.2 & 6338 & 14.5\% & 74.3 & \textbf{38.8\%} \\
CGRS & 94.5 & 3235 & 28.9\% & 93.3 & 5076 & 29.4\% & 70.0 & 8662 & 23.5\% & 65.2 & 5953 & 19.7\% & \textbf{80.8} & 25.4\% \\
\rowcolor{gray!15}\method & 94.6 & 2992 & 34.3\% & 93.3 & 5070 & 29.5\% & 70.0 & 9426 & 16.7\% & 64.1 & 5027 & 32.2\% & \underline{80.5} & \underline{28.2\%} \\
\bottomrule
\end{tabular}
}
\caption{Comparison across models of different scales and multiple methods on four benchmarks. Acc (\%) denotes accuracy, \#Tok denotes average response length in tokens, and LR denotes length reduction ratio. Higher Acc and LR, and lower \#Tok indicate better performance. Best average results are in \textbf{bold}, second best are \underline{underlined}.}
\label{tab:main}
\end{table*}

\section{Experiments}

\subsection{Experimental Setup}

\paragraph{Benchmarks and Metrics.}
We conduct experiments on a set of mathematical and scientific reasoning benchmarks.
For mathematical reasoning, we consider three datasets: MATH500~\cite{lightman2023let}, a collection of 500 multi-step problems spanning algebra, geometry, and probability; AMC23~\cite{amc23}, which contains 40 problems from the 2023 American Mathematics Competitions; and AIME24~\cite{aime}, consisting of 30 challenging problems from the 2024 American Invitational Mathematics Examination.
In addition, we evaluate scientific reasoning performance on GPQA Diamond~\cite{rein2024gpqa} (GPQA-D), which includes 198 graduate-level multiple-choice questions across biology, chemistry, and physics. We report three evaluation metrics: \textit{Accuracy} (\textbf{Acc}), the average number of generated tokens (\textbf{\#Tok}), and the \textit{Length Reduction Rate} (\textbf{LR}).

\paragraph{Baselines.}
We compare our method with the following baselines:
(i) \textbf{Vanilla}, which performs standard decoding without any intervention;
\textbf{NoThinking}~\cite{ma2025reasoning}, which skips the reasoning process entirely.
(ii) Prompt-guided methods, including \textbf{TALE}~\cite{han2025token}, which prompts the model to solve problems within token budgets.
(iii) Output-based methods, including \textbf{Dynasor}~\cite{fu2024efficiently}, which periodically requests intermediate answers at fixed token intervals and exits early if multiple consecutive answers match;
\textbf{DEER}~\cite{yang2025dynamic}, which dynamically truncates chain-of-thought generation by detecting high-confidence intermediate answers;
and \textbf{CGRS}~\cite{huang2025efficient}, which adjusts the ratio of reflection tokens according to answer confidence.

\paragraph{Implementation Details.}

For calibration, we randomly sample a subset from the training split of MATH~\cite{hendrycks2021measuring} and generate reasoning traces using greedy decoding. Attribution scores (Eq.~\ref{eq:importance}) are computed for FFN neurons at the termination timestep. Temporal Filtering and cross-sample consistency filtering are then applied, resulting in a compact neuron set $\mathcal{S}^*$, with the exact size depending on the model. During inference, the intervention is triggered based on similarity and magnitude criteria. Each experiment is repeated three times and the average results are reported. We use the vLLM framework~\cite{kwon2023efficient} for inference acceleration. All decoding is performed with temperature 0.6 and top-$p$ 0.95. Hyperparameter settings are provided in Appendix~\ref{appendix:hyperparams}.

\subsection{Main Results}

Table~\ref{tab:main} compares \textsc{\method} with representative baselines across four reasoning benchmarks and four models spanning different architectures and scales. Overall, \textsc{\method} consistently achieves around 22\% to 28\% average token reduction while maintaining comparable or slightly improved accuracy relative to Vanilla decoding, yielding a stable efficiency--accuracy trade-off. Additional results on smaller-scale models are provided in Appendix~\ref{appendix:smaller}.

In contrast, output-based early-exit methods that rely on intermediate answer confidence estimation exhibit substantial variance across models. For example, DEER achieves aggressive compression on stronger and better-calibrated models such as Qwen3, reducing generation length by 41.7\% on average, but often at the cost of accuracy. On AIME24 with Qwen3-8B, DEER drops accuracy from 61.1\% to 45.6\%. Similar behavior is observed on Llama-based models, where both DEER and CGRS suffer notable accuracy degradation while achieving only marginal length reduction. On GPQA-D with Llama-8B, DEER and CGRS reduce accuracy by 7.5 and 9.7 points respectively, with length reduction below 6\%, indicating premature or ineffective exits. These results suggest that output-based heuristic exit methods are highly sensitive to model calibration.

In comparison, \textsc{\method} more consistently preserves accuracy while maintaining non-trivial compression. Importantly, the difference is not only attributable to calibration sensitivity but also to the intervention strategy. Methods such as DEER and Dynasor employ direct and aggressive termination, immediately halting decoding once their criteria are met, which increases the risk of irreversible errors. By contrast, \textsc{\method} adopts a graded exit mechanism, preserving more opportunities for correcting erroneous reasoning paths before termination. Overall, these results suggest that internal neuron activation dynamics provide a more reliable proxy for early reasoning exit than surface-level signal.

\subsection{Inference Latency.}
\begin{figure}[h!]
    \centering
    \includegraphics[width=\columnwidth]{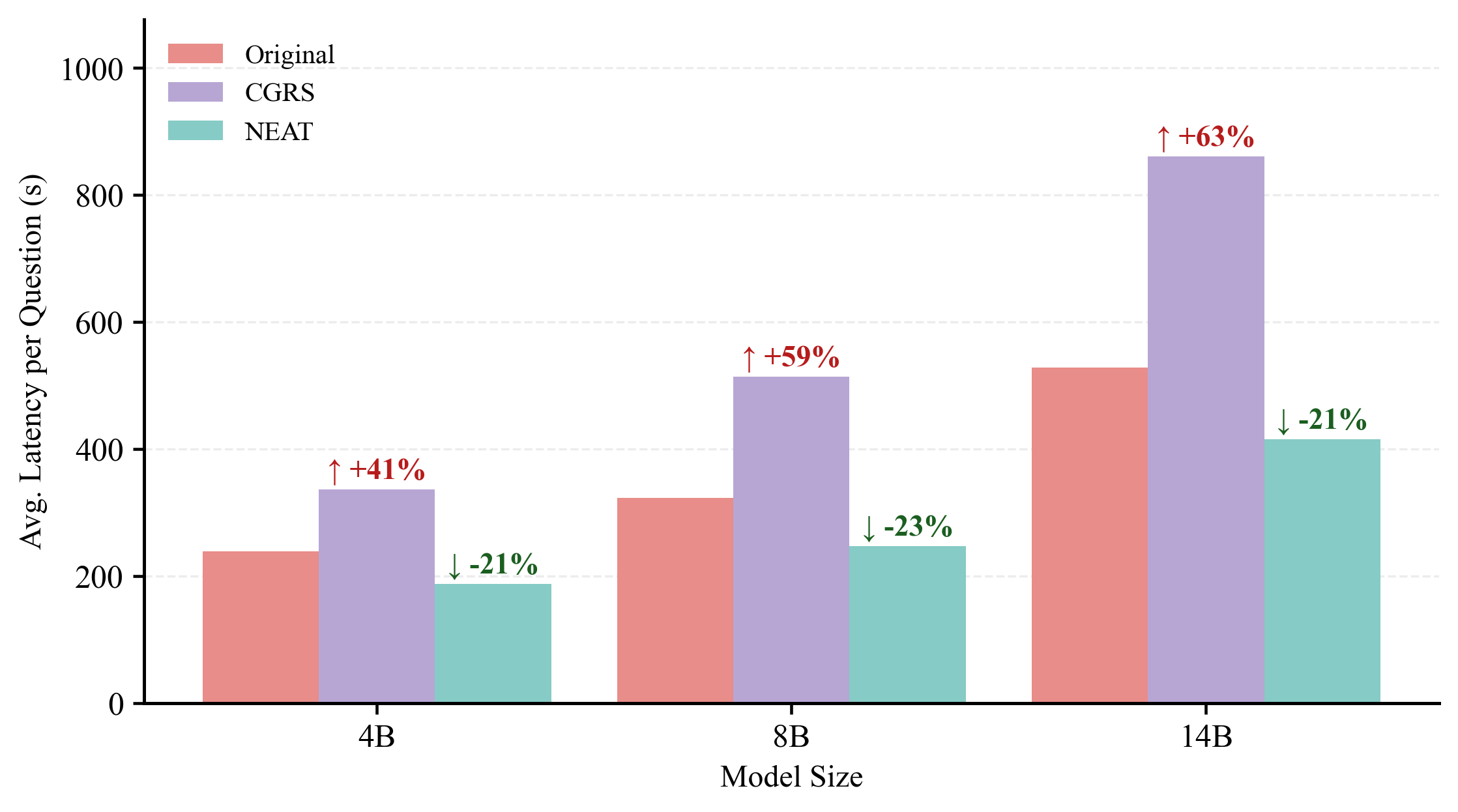}
    \caption{Inference latency (seconds per question) on AIME2024 across Qwen3-series models of different scales.}
    \label{fig:latency}
\end{figure}
Figure~\ref{fig:latency} compares inference latency on AIME2024 across Qwen3-series models of different scales. Although output-based heuristic methods aim to reduce generation length, they can introduce substantial runtime overhead. As shown in Figure~\ref{fig:latency}, CGRS consistently incurs higher latency than Vanilla decoding, with increases ranging from 41\% to 63\% across model sizes, despite reducing token count. This overhead arises from additional computation required for confidence estimation during decoding. In contrast, \textsc{\method} achieves consistent latency reductions of approximately 21\% to 23\% across all model scales. Since our approach relies solely on monitoring internal neuron activations and does not require extra forward passes or parallel sampling, the reduction in generation length directly translates into wall-clock speedups. These results demonstrate that internal activation-based intervention provides a more computationally efficient alternative to output-based early-exit methods.

\subsection{Ablation Study}

    \begin{table}[t]
    \centering
    \small
    \begin{tabular}{lcc}
    \toprule
    \textbf{Method} & Accuracy ($\uparrow$) & \#Tok ($\downarrow$) \\
    \midrule
    Vanilla & 92.2 & 3743 \\
    \method & 92.2 & 2914 \\
    \midrule
    \multicolumn{3}{l}{\textit{Suppression-only}} \\
    \quad $\tau_{\text{sup}}=0.6$ & \textbf{93.4} & 3485 \\
    \quad $\tau_{\text{sup}}=0.4^{*}$ & 93.2 & 3058 \\
    \quad $\tau_{\text{sup}}=0.2$ & 89.6 & 2763 \\
    \midrule
    \multicolumn{3}{l}{\textit{Exit-only}} \\
    \quad $\tau_{\text{sim}}=0.8$ & 92.8 & 3604 \\
\quad $\tau_{\text{sim}}=0.6^{*}$ & 92.6 & 3232 \\
    \quad $\tau_{\text{sim}}=0.4$ & 87.2 & \textbf{2287} \\
    \bottomrule
    \end{tabular}
    \caption{Ablation of inference-time intervention components for \textsc{\method} on MATH500 using DeepSeek-R1-Distill-Qwen-7B. $^{*}$ denotes the default configuration used in our main experiments.}
    \label{tab:ablation_exit}
    \end{table}

\begin{figure}[h!]
    \centering
    \includegraphics[width=0.45\textwidth]{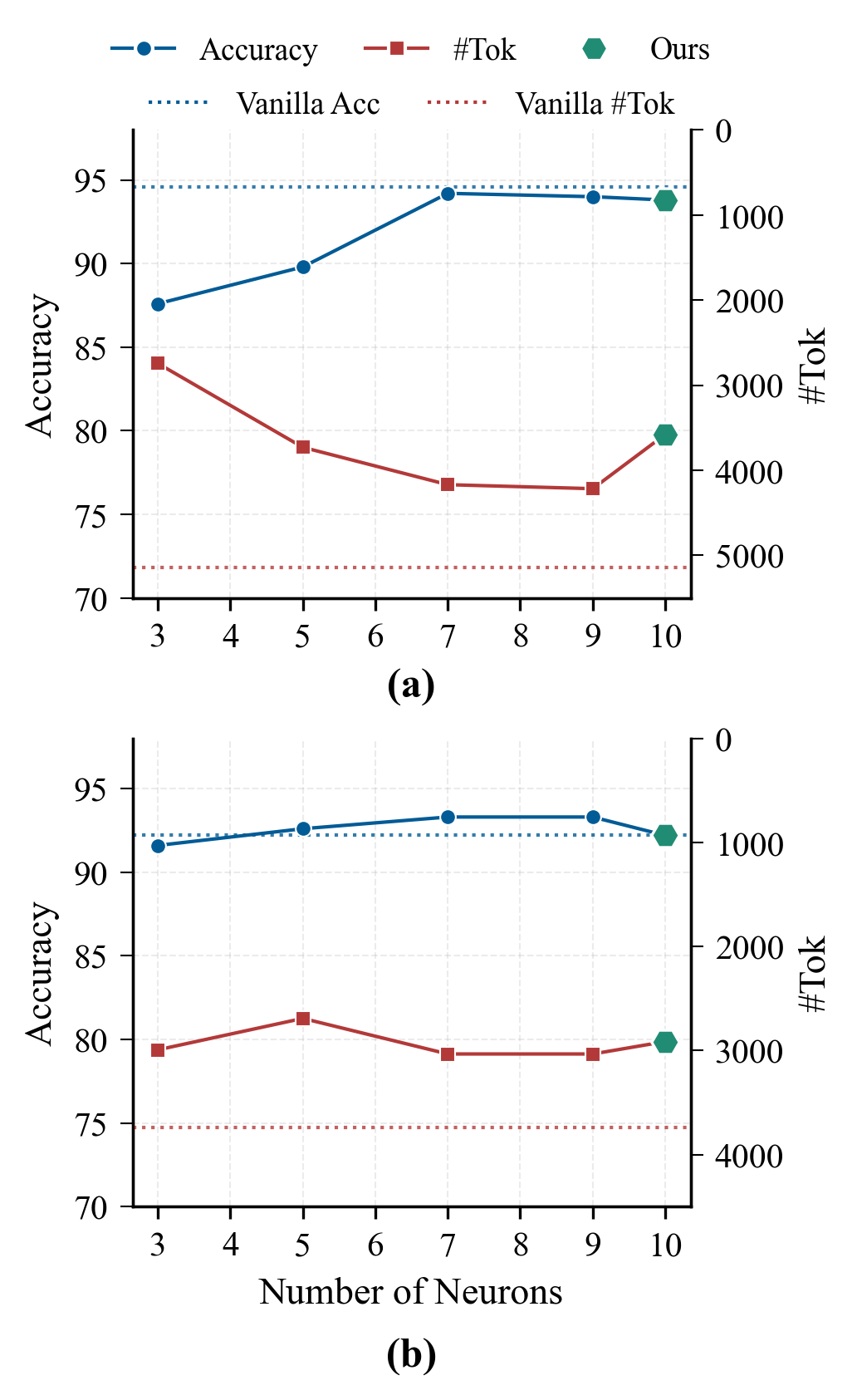}
    \caption{Sensitivity analysis with respect to the number of monitored neurons in $\mathcal{S}^*$. Blue solid lines denote accuracy (left y-axis), and red dashed lines denote average response length (\#Tok, right y-axis), where the right axis is inverted. Horizontal dotted lines indicate the Vanilla baseline. \textbf{(a)} Qwen3-8B. \textbf{(b)} DeepSeek-R1-Distill-Qwen-7B.}

    \label{fig:ablation}
\end{figure}

\paragraph{Effect of Number of Monitored Neurons.}
Figure~\ref{fig:ablation} shows the impact of the number of monitored neurons in $\mathcal{S}^*$. Sensitivity to this choice varies across models.
For Qwen3-8B (Figure~\ref{fig:ablation}(a)), monitoring too few neurons noticeably degrades accuracy, while accuracy quickly recovers and stabilizes as more neurons are included, with diminishing returns beyond a small threshold.
In contrast, DeepSeek-R1-Distill-Qwen-7B (Figure~\ref{fig:ablation}(b)) shows much lower sensitivity, maintaining stable accuracy across a wide range of neuron counts. This robustness arises from the combination of neuron attribution and temporal filtering. Neurons are ranked by attribution scores (Eq.~\ref{eq:importance}) and further required to exhibit concentrated late-stage activation, so additional neurons beyond the most relevant ones tend to have weaker or noisier signals and provide limited benefit. Token length varies only moderately with different $k$, indicating that \textsc{\method} largely preserves efficiency as long as a sufficient subset of exit-associated neurons is monitored.

\paragraph{Intervention Components.}
Table~\ref{tab:ablation_exit} studies inference-time intervention components by separately ablating suppression and exit mechanisms of \textsc{\method} on MATH500 with DeepSeek-R1-Distill-Qwen-7B.
In the \textit{Suppression-only} setting, the suppression threshold $\tau_{\text{sup}}$ controls intervention strength without forcing early termination. Moderate suppression improves accuracy while reducing token usage compared to Vanilla decoding, whereas overly strong suppression significantly harms accuracy, suggesting that excessive interference can disrupt valid reasoning even without hard exits.
In the \textit{Exit-only} setting, the similarity threshold $\tau_{\text{sim}}$ determines early termination frequency. Lower thresholds yield greater token savings but cause notable accuracy degradation due to premature exits, while higher thresholds better preserve accuracy but offer limited efficiency gains.
Overall, suppression provides smoother and more robust control, while exit decisions are more sensitive to threshold selection. Combining both mechanisms enables \textsc{\method} to achieve a better balance between accuracy and inference efficiency, motivating the default configuration used in our main experiments.

\subsection{Further Analysis}
\label{sec:analysis}

\begin{table}[t]
\centering
\resizebox{\columnwidth}{!}{%
\begin{tabular}{lcc}
\toprule
\textbf{Method} & Accuracy ($\uparrow$) & \#Tok ($\downarrow$) \\
\midrule
Vanilla       & 92.2 & 3743 \\
NoThinking    & 80.9 & 1173 \\
\midrule
\multicolumn{3}{l}{\textit{Logit-based}} \\
\quad Wait~\cite{wang2025wait} & 87.8 & 2829 \\
\quad \texttt{</think>}~\cite{liu2025answer} & 88.8 & 2373 \\
\quad Answer~\cite{yang2025dynamic} & \textbf{92.0} & 2844 \\
\midrule
\multicolumn{3}{l}{\textit{Hidden-state-based}} \\
\quad Similarity & 80.6 & 1301 \\
\quad Probe~\cite{zhang2025reasoning} & 82.0 & 1855 \\
\midrule
\multicolumn{3}{l}{\textit{Neuron-based}} \\
\quad Random & 77.8 & 1562 \\
\quad Top Activated & 88.8 & 2458 \\
\quad Ours & \textbf{92.2} & 2914 \\
\bottomrule
\end{tabular}}
\caption{Comparison of different exit signals on MATH500 using DeepSeek-R1-Distill-Qwen-7B.}
\label{tab:signal}
\end{table}

\paragraph{Comparison of Exit Signals.}
Table~\ref{tab:signal} compares different early-exit signals on MATH500, including surface-level logits, hidden states, and neuron-level activations. Implementation details are provided in Appendix~\ref{appendix:baselines}.

Among surface-level signals, answer-based confidence outperforms simple logit heuristics (e.g., \texttt{</think>} or hesitation tokens), but still lags behind Vanilla decoding in accuracy and often incurs additional test-time overhead. Moreover, directly suppressing explicit reasoning degrades performance, suggesting that reflective reasoning remains important for difficult problems.

Hidden-state-based signals are less reliable. Although hidden representations contain confidence-related information, similarity- and probing-based methods achieve large length reductions at the cost of substantial accuracy loss, indicating that they often trigger premature exits.

In contrast, neuron-level signals provide a better balance between accuracy and efficiency. Random or top-activated neurons fail to serve as reliable exit indicators, implying that termination cues are sparse and not solely determined by activation magnitude. By identifying exit-associated neurons, \textsc{\method} achieves near-Vanilla accuracy while reducing generation length by over 20\%, with significantly lower inference overhead than answer-based signals.

\paragraph{Case Study.}
Figure~\ref{fig:case_math500} presents a representative example illustrating how NEAT avoids redundant reasoning after a correct solution has been reached. In this case, the model correctly determines that the two direction vectors are perpendicular, yielding an angle of $90^\circ$. After the correct answer is obtained, the activation patterns of exit-associated neurons progressively align with the reference regime, which triggers suppression and an early exit from further decoding. As a result, NEAT terminates generation after 1,358 tokens while preserving the correct answer and stable reasoning outcome. By contrast, Vanilla decoding continues for an additional 840 tokens, repeatedly revisiting the same conclusion through reflective and confirmatory reasoning steps. Despite the absence of new information or changes to the final answer, the model engages in extended verification and re-computation, leading to substantial redundant continuation without improving correctness.

\begin{figure}[t]
\centering
\begin{tcolorbox}[
  enhanced,
  colback=orange!4,
  colframe=orange!60!black,
  boxrule=0.4pt,
  arc=1.5pt,
  left=2pt,right=2pt,top=2pt,bottom=2pt,
  boxsep=1pt,
  title=An example from MATH500,
  colbacktitle=orange!35,
  coltitle=black,
  fonttitle=\bfseries\small,
  fontupper=\small,
]
\raggedright
\setlength{\parskip}{0pt}\setlength{\parindent}{0pt}

\textbf{Problem:}
The set of points $(x,y,z)$ that satisfy 2x = 3y = -zis a line. The set of points $(x,y,z)$ that satisfy 6x = -y = -4z is another line. Find the angle between these lines, in degrees.\\
\textbf{Ground Truth:} $90^\circ$\\
{\color{black}\rule{\linewidth}{0.3pt}}\\[-1pt]
\textbf{Response:}\\
Okay, so I have this problem where I need to find the angle between two lines in 3D space...\\
3*2 = 6,2*(-12) = -24,-6*(-3) = 18\\
Adding them up: 6 - 24 + 18 = 0 \textcolor{green!50!black}{[Suppression: Sim:0.48]}\\
That means the vectors are perpendicular,so the angle between them is 90 degrees. \\
I double-check my calculations because that seems a bit surprising.
First...
Yes, that's correct. So the angle between them is 90 degrees.\textcolor{green!80!black}{[Exit: Sim:0.68]}\\
\textcolor{gray!60}{Total 1358 token}
{\color{black}\rule{\linewidth}{0.3pt}}\\[-1pt]
\textbf{Vanilla (Continue):}\\
\textcolor{red!95!black}{But wait[Reflection]}, the problem says to round the answer to the direction of the direction vector of the second line. Hmm, I'm not sure what that means. ...\\
So, I think the answer is 90 degrees.\\
But just to be thorough, let me compute the angle using the formula.\textcolor{orange!90!black}{[Overthinking]}
...So, the angle between the two lines is 90 degrees.
\\
\textcolor{gray!60}{Total 2198 tokens}
\end{tcolorbox}

\caption{Case study illustrating overthinking in Vanilla decoding and how NEAT avoids redundant continuation.}
\label{fig:case_math500}
\end{figure}

\section{Conclusion}
We study the problem of overthinking in large reasoning models and propose \textsc{\method}, a training-free neuron-based framework for early reasoning exit. By identifying exit-associated neurons and monitoring their activation dynamics during inference, \textsc{\method} performs early exit or suppresses reflection to reduce redundant reasoning steps. Experiments across multiple benchmarks and model architectures show that \textsc{\method} consistently reduces generation length while maintaining accuracy comparable to Vanilla decoding. Overall, our results demonstrate that neuron-level activation dynamics provide an effective and efficient signal for controlling reasoning progression without external supervision or parallel rollouts.

\section*{Limitations}
Our approach assumes access to internal model activations, which may not be available in all deployment scenarios such as API-only settings, and thus limits applicability to models with full access. In addition, our experiments are conducted on models up to 14B parameters, and it remains an open question whether the same neuron identification and calibration procedure generalizes to substantially larger models. Finally, while activation monitoring introduces minimal overhead in our current implementation, integrating such mechanisms into highly optimized inference engines may require additional engineering effort. We leave these directions for future work.

\section*{Ethics Statement}
This work focuses on improving inference efficiency in large reasoning models through internal activation analysis. It does not involve human subjects, personal data, or user-generated content. We do not anticipate direct negative societal impacts from the proposed method. As with other techniques that improve model efficiency, responsible deployment should follow existing best practices for large language models.



\bibliography{custom}
\newpage
\appendix

\section{More Implementation Details}


\subsection{Reflection Token List}

Table~\ref{tab:reflect_tokens} presents the complete list of reflection tokens used for soft suppression. We include both variants with and without a leading space.
\begin{table}[h]
\centering
\label{tab:reflection_tokens}
\begin{tabular}{ll}
\toprule
\textbf{Token} & \textbf{Variant} \\
\midrule
\texttt{Wait} & \texttt{" Wait"} \\
\texttt{Hmm} & \texttt{" Hmm"} \\
\texttt{But} & \texttt{" But"} \\
\texttt{Alternatively} & \texttt{" Alternatively"} \\
\texttt{However} & \texttt{" However"} \\
\bottomrule
\end{tabular}
\caption{Reflection tokens used for soft suppression.}\label{tab:reflect_tokens}
\end{table}

\subsection{Latency Experiment Details}
\label{appendix:latency}
For each model, we conduct latency experiments on a single NVIDIA A6000 GPU (48GB) using the vLLM framework~\citep{kwon2023efficient}. GPU memory utilization is set to 0.9, and the maximum number of generated tokens is set to 16,384. We report the average inference time per question in seconds.







\subsection{Exit Signal Implementation Details}
\label{appendix:baselines}

\paragraph{Output-based Methods.}
\textbf{Answer}~\cite{huang2025efficient}: We use the answer-based early-exit method proposed in CGRS~\cite{huang2025efficient}.  
\textbf{Wait}~\cite{wang2025wait}: We suppress reflection tokens (e.g., Wait) during generation.  
\textbf{\texttt{</think>}}~\cite{liu2025answer}: We increase the logits of the termination token to encourage earlier stopping.

\paragraph{Hidden-state-based Methods.}
\textbf{Similarity}: We compute the cosine similarity between the current last-layer hidden state and a reference state, which is obtained by averaging hidden states over calibration samples. The exit threshold is set to 0.4.  
\textbf{Probe}~\cite{zhang2025reasoning}: We use a pretrained linear probe on hidden states to predict reasoning correctness. The exit threshold is set to 0.85.

\paragraph{Neuron-based Methods.}
\textbf{Random}: We randomly select neurons at the termination position, using the same number of neurons and layer distribution as NEAT. The exit threshold is set to 0.6.  
\textbf{Top-Activated}: We select neurons with the highest activation values at the termination position, using the same number of neurons and layer distribution as NEAT. The exit threshold is set to 0.6.

\subsection{Prompt Template}
\label{appendix:prompt}
Figure~\ref{fig:prompt-math} shows the prompt used for MATH / AMC / AIME.
Figure~\ref{fig:prompt-gpqa} shows the prompt we use for dataset GPQA-D. 
\FloatBarrier
\begin{figure}[!t]
\centering
\begin{promptbox}{Prompt Template (MATH / AMC / AIME)}
\texttt{\{problem\}}\par\vspace{6pt}
Please reason step by step, and put your final answer within
\texttt{\textbackslash boxed\{\}}.
\end{promptbox}
\caption{Prompt template used for MATH500, AMC23, and AIME24.}
\label{fig:prompt-math}
\end{figure}

\begin{figure}[!t]
\centering
\begin{promptbox}{Prompt Template (GPQA-D)}
\texttt{\{problem\}}\par\vspace{6pt}
Please reason step by step and output only the choice letter
within \texttt{\textbackslash boxed\{\}}.
\end{promptbox}
\caption{Prompt template used for GPQA-D.}
\label{fig:prompt-gpqa}
\end{figure}

\subsection{Hyperparameter Settings}
\label{appendix:hyperparams}
We evaluate our method on six models spanning different scales and architectures: DeepSeek-R1-Distill-Qwen-1.5B/7B, DeepSeek-R1-Distill-Llama-8B, and Qwen3-4B/8B/14B. The main hyperparameters we use are shown in Table~\ref{tab:hyperparams}. The termination token $w_{\text{end}}$ is set to \texttt{</think>} for DeepSeek-R1-Distill-Qwen-1.5B/7B, and \texttt{**} for all other models.

\begin{table}[!h]
\centering
\begin{tabular}{c c}
\toprule
\textbf{Param.} & \textbf{Value} \\
\midrule
$|\mathcal{D}_{\text{cal}}|$ & 20 \\
$k$ & 300 \\
$\tau_{\text{com}}$ & 0.6 \\
$\tau_{\text{cons}}$ & 0.6 \\
$\tau_{\text{sim}}$ & 0.6 \\
$\tau_{\text{sup}}$ & 0.4 \\
$\tau_{\text{mag}}$ & 0.2 \\
$w_{\text{end}}$ & $\{\texttt{</think>}, \texttt{**}\}$ \\
\bottomrule
\end{tabular}
\caption{Hyperparameters used in our experiments.}
\label{tab:hyperparams}
\end{table}


\section{Results on Smaller Models}\label{appendix:smaller}

\begin{table*}[!t]
\centering
\small

\resizebox{\textwidth}{!}{%
\begin{tabular}{lccc ccc ccc ccc cc}
\toprule
\multirow{2}{*}{\textbf{Method}} 
& \multicolumn{3}{c}{\textbf{MATH500}} 
& \multicolumn{3}{c}{\textbf{AMC23}} 
& \multicolumn{3}{c}{\textbf{AIME24}} 
& \multicolumn{3}{c}{\textbf{GPQA-D}} 
& \multicolumn{2}{c}{\textbf{AVG}} \\
\cmidrule(lr){2-4} \cmidrule(lr){5-7} \cmidrule(lr){8-10} \cmidrule(lr){11-13} \cmidrule(lr){14-15}
 & Acc$\uparrow$ & \#Tok$\downarrow$ & LR$\uparrow$
 & Acc$\uparrow$ & \#Tok$\downarrow$ & LR$\uparrow$
 & Acc$\uparrow$ & \#Tok$\downarrow$ & LR$\uparrow$
 & Acc$\uparrow$ & \#Tok$\downarrow$ & LR$\uparrow$
 & Acc$\uparrow$ & LR$\uparrow$ \\
\midrule
\multicolumn{15}{c}{\textit{DeepSeek-R1-Distill-Qwen-1.5B}} \\
\midrule
Vanilla & 84.8 & 4687 & -- & 72.5 & 6789 & -- & 36.6 & 10577 & -- & 36.6 & 7742 & -- & 57.6 & -- \\
NoThinking & 71.9 & 1759 & 62.5\% & 63.3 & 3725 & 45.1\% & 18.9 & 7831 & 26.0\% & 27.6 & 2202 & 71.6\% & 45.4 & 51.3\% \\\hdashline
TALE & 82.9 & 4387 & 6.4\% & 70.0 & 7486 & -10.3\% & 31.0 & 11701 & -10.6\% & 32.3 & 6886 & 11.1\% & \underline{54.1} & -0.9\% \\
Dynasor & 79.3 & 2678 & 42.9\% & 70.8 & 5661 & 16.6\% & 25.6 & 9745 & 7.9\% & 34.5 & 2998 & 61.3\% & 52.6 & 32.2\% \\
DEER & 79.6 & 2665 & 43.2\% & 67.5 & 5097 & 25.0\% & 28.3 & 9564 & 9.6\% & 34.8 & 6900 & 10.9\% & 52.6 & 22.2\% \\
CGRS & 75.5 & 2174 & 53.6\% & 62.5 & 4180 & 38.4\% & 26.7 & 8082 & 23.6\% & 30.0 & 3257 & 57.9\% & 48.7 & \textbf{43.4\%} \\
\rowcolor{gray!15}\method & 82.7 & 3223 & 31.2\% & 72.5 & 4580 & 32.5\% & 31.1 & 8956 & 15.3\% & 35.4 & 6787 & 12.3\% & \textbf{55.0} & \underline{22.8\%} \\
\midrule
\multicolumn{15}{c}{\textit{Qwen3-4B}} \\
\midrule
Vanilla & 92.7 & 4796 & -- & 91.7 & 7449 & -- & 60.0 & 11449 & -- & 54.0 & 8112 & -- & 74.6 & -- \\
NoThinking & 84.9 & 988 & 79.4\% & 70.0 & 1710 & 77.0\% & 24.4 & 4504 & 60.7\% & 47.5 & 1471 & 82.1\% & 56.7 & 74.8\% \\\hdashline
TALE & 89.1 & 2657 & 44.6\% & 86.7 & 5107 & 31.4\% & 48.9 & 9727 & 15.0\% & 36.2 & 4938 & 39.1\% & 65.2 & 32.5\% \\
Dynasor & 90.1 & 3877 & 19.2\% & 86.7 & 6233 & 16.3\% & 54.3 & 9912 & 13.4\% & 50.5 & 4398 & 45.8\% & 70.4 & 23.7\% \\
DEER & 87.0 & 1854 & 61.3\% & 80.0 & 3231 & 56.6\% & 50.0 & 6873 & 40.0\% & 54.9 & 7033 & 13.3\% & 68.0 & \textbf{42.8\%} \\
CGRS & 91.3 & 2704 & 43.6\% & 86.7 & 4351 & 41.6\% & 56.7 & 7893 & 31.1\% & 55.2 & 5229 & 35.5\% & \underline{72.5} & \underline{37.9\%} \\
\rowcolor{gray!15}\method & 94.0 & 3415 & 28.8\% & 89.1 & 5219 & 29.9\% & 57.4 & 9174 & 19.9\% & 55.5 & 6625 & 18.3\% & \textbf{74.0} & 24.2\% \\
\bottomrule
\end{tabular}%
}
\caption{Comparison of methods on smaller-scale models: DeepSeek-R1-Distill-Qwen-1.5B and Qwen3-4B.}
\label{tab:smaller_models}
\end{table*}

Table~\ref{tab:smaller_models} presents results on smaller-scale models with 1.5B and 4B parameters. Compared to larger models, smaller models are generally more sensitive to premature termination, making early-exit strategies more challenging. Despite this, \textsc{\method} consistently preserves accuracy close to the Vanilla baseline while achieving meaningful length reduction across benchmarks. In contrast, output-based methods often obtain higher compression at the cost of substantial accuracy degradation, particularly on the 1.5B model. These results indicate that neuron-level exit signals remain robust across model scales and enable a stable efficiency–accuracy trade-off even on smaller models.

\end{document}